\pdfoutput=1

\documentclass[11pt]{article}

\usepackage[hyperref]{emnlp2021}

\usepackage{times}
\usepackage{latexsym}

\usepackage[T1]{fontenc}

\usepackage[utf8]{inputenc}

\usepackage{microtype}

\usepackage{times}
\usepackage{latexsym}
\usepackage{booktabs}
\usepackage{xspace}
\usepackage{graphicx}
\usepackage{enumitem}
\usepackage{mathtools}
\usepackage{subcaption}
\usepackage{xcolor}
\usepackage{float}
\usepackage{comment}
\usepackage{bibunits} %

\definecolor{posterV}{RGB}{89,89,89}
\definecolor{posterC}{RGB}{255,172,63}
\definecolor{posterR}{RGB}{109,158,235}

\usepackage{microtype}

\title{\papertitle}

\author{
Jack Hessel$^{\dagger}$~~~Ari Holtzman$^{\ddagger}$~~~Maxwell Forbes$^{\ddagger}$~~~Ronan Le Bras$^{\dagger}$~~~Yejin Choi$^{\dagger\ddagger}$ \\
$^\dagger$Allen Institute for AI\\
$^\ddagger$Paul G. Allen School of Computer Science \& Engineering, University of Washington\\
    { \tt\small \{jackh,ronanlb\}@allenai.org \{ahai,mbforbes,yejin\}@cs.washington.edu} \vspace{0.5em}
    }

\date{}

\newcommand{\metric}[1]{{\small #1}\xspace}

\newcommand{\bleu}{\metric{BLEU-4}}
\newcommand{\bleufour}{\metric{BLEU-4}}
\newcommand{\bleuone}{\metric{BLEU-1}}
\newcommand{\meteor}{\metric{METEOR}}
\newcommand{\rouge}{\metric{ROUGE-L}}
\newcommand{\cider}{\metric{CIDEr}}
\newcommand{\spice}{\metric{SPICE}}
\newcommand{\tiger}{\metric{TIGEr}}
\newcommand{\leic}{\metric{LEIC}}
\newcommand{\clipscore}{\metric{CLIP-S}}
\newcommand{\clipscorenorefs}{\clipscore (no refs)}
\newcommand{\clipscorelong}{\texttt{CLIPScore}\xspace}
\newcommand{\alttext}{alt-text\xspace}

\newcommand{\bertscorerobertaf}{\metric{BERT-S (RoBERTa-F)}}
\newcommand{\bertscore}{\metric{BERT-S}}
\newcommand{\bertscoretbr}{\metric{BERT-S++}}

\newcommand{\vilbertscore}{\metric{ViLBERTScore-F}}
\newcommand{\refclipscore}{\metric{RefCLIP-S}}
\newcommand{\refclipscorelong}{\texttt{RefCLIPScore}\xspace}

\newcommand{\reqref}{reference-based\xspace}
\newcommand{\Reqref}{Reference-based\xspace}

\newcommand{\papertitle}{\clipscorelong:\\
A Reference-free Evaluation Metric for Image Captioning}

\definecolor{cborange}{HTML}{e66101}
\definecolor{cbpurple}{HTML}{5e3c99}

\begin{document}

\maketitle

\begin{bibunit}[acl_natbib]

\begin{abstract}

Image captioning has conventionally relied on \emph{\reqref} automatic evaluations, where machine captions are compared against captions written by humans. This is in
contrast to the \emph{reference-free} manner in which humans assess caption quality.

In this paper, we report the surprising empirical finding that CLIP \cite{radford2learning}, a cross-modal model pretrained on 400M image+caption pairs from the web, can be used for robust automatic evaluation of image captioning without the need for references. Experiments spanning several corpora demonstrate that our new reference-free metric, \clipscorelong, achieves the highest correlation with human judgements, outperforming existing \reqref metrics like \cider and \spice.
Information gain experiments demonstrate that \clipscorelong, with its tight focus on \emph{image--text} compatibility,
is complementary to existing \reqref metrics that emphasize \emph{text--text} similarities. Thus, we also present a reference-augmented version, \refclipscorelong, which achieves even higher correlation.
Beyond literal description tasks, several case studies reveal domains where \clipscorelong performs well (clip-art images, alt-text rating), but also where it is relatively weaker in comparison to \reqref metrics, e.g., news captions that require richer contextual knowledge.

\end{abstract}

\section{Introduction}

\begin{figure}[t!]
    \centering
    \includegraphics[width=.95\linewidth]{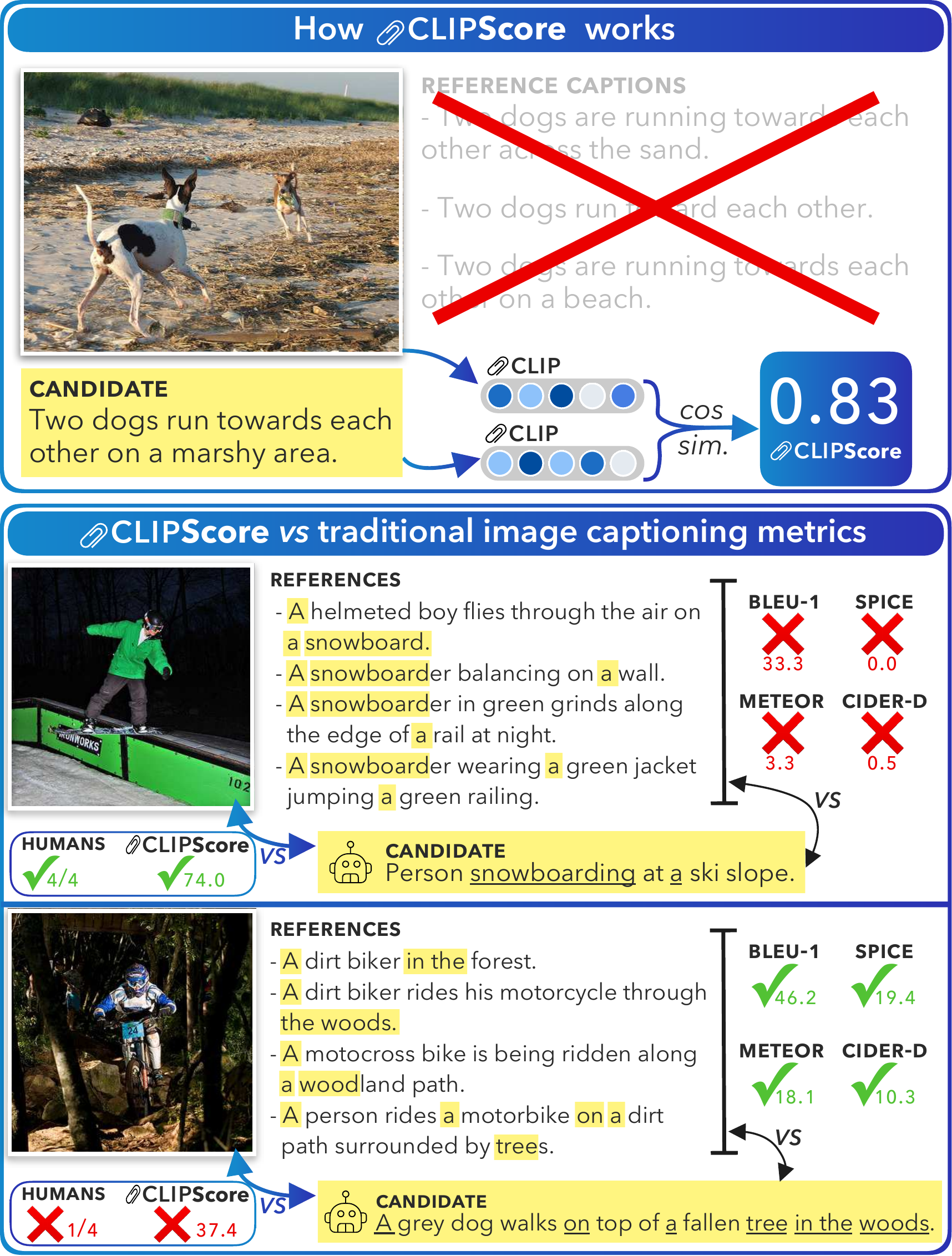}
    \caption{
\textbf{Top:} \clipscorelong uses CLIP to assess image-caption compatibility \emph{without} using references, just like humans. \textbf{Bottom:} This frees \clipscorelong from the well-known shortcomings of $n$-gram matching metrics, which disfavor good captions with new words (top) and favor \emph{any} captions with familiar words (bottom). \small{Attribution: Paperclip, robot icons by Hasanudin, Adiyogi (resp.) from the Noun Project.}
    }
    \label{fig:failure_cases}
\end{figure}

For most text generation tasks, \reqref  $n$-gram overlap methods are still the dominant means of automatic evaluation. For image caption generation, recent \reqref metrics have sought to transcend overlap by considering richer models of reference-candidate similarity: e.g., approximate scene graphs \cite{anderson2016spice}, allowing \reqref methods to incorporate the image \cite{jiang2019tiger,lee2020vilbertscore}. But, references can be expensive to collect and comparing against even \emph{multiple} human-authored captions for each image is often insufficient (see \autoref{fig:failure_cases}). As a result, for many corpora, a significant gap remains between \reqref scoring and human quality judgments.\footnote{See \newcite{elliott2014comparing} and \newcite{kilickaya2016re} for thorough comparisons of caption generation metrics.}

Should we need references for the evaluation of image captions? After all, when humans assess the appropriateness of an image caption, we do so just by looking at the image and reading the candidate's text.

A recent trend in machine translation serves as inspiration: there, a key hurdle for reference-free evaluation (sometimes called \emph{quality estimation}) has been estimating cross-lingual similarity between source+candidate pairs \cite{blatz2004confidence,specia2010machine,mehdad2012match,specia2018machine}. But recent work \cite{lo2019yisi,yankovskaya2019quality,zhao-etal-2020-limitations} has improved correlation with human judgment not by gathering more monolingual references, but instead by utilizing cross-lingual representations learned by large-scale, pre-trained, multilingual models e.g., LASER \cite{artetxe2019massively} or M-BERT \cite{devlin2018bert}. \footnote{\newcite{wang2019cross}, \newcite{pires2019multilingual}, and \newcite{Wu:2019rw} explore how M-BERT learns and utilizes cross-lingual information.} 

We hypothesize that the relationships learned by pretrained vision+language models (e.g., ALIGN \cite{jia2021scaling} and CLIP \cite{radford2learning}) could similarly support reference-free evaluation in the image captioning case. %
Indeed, they can: we show that a relatively direct application of CLIP to (image, generated caption) pairs results in surprisingly high correlation with human judgments on a suite of standard image description benchmarks (e.g., MSCOCO \cite{lin2014microsoft}). We call this process \clipscorelong (abbreviated to \clipscore). Beyond direct correlation with human judgments, an information gain analysis reveals that \clipscore is complementary both to commonly reported metrics (like \bleu, \spice, and \cider) and to newly proposed \reqref metrics (e.g., \vilbertscore \cite{lee2020vilbertscore}).

We additionally (1) propose a reference-augmented version of \clipscorelong, \refclipscorelong, that achieves even higher human correlation, (2) verify that \clipscore is sensitive to adversarially constructed image captions, where one noun-phrase has been swapped for a plausible (but incorrect) distractor; and (3) construct a corpus of images that have never been posted publicly online to verify that \clipscore is able to reconstruct human judgments on never-before-seen images.

Finally, we assess \clipscore in the context of four case studies that diverge from context-free, literal photograph description. In two cases, \clipscore works well: it achieves high correlation with \alttext quality rating on Twitter, and demonstrates surprising capacity to reason about clipart images+captions. For news caption generation, \reqref methods correlate best with human judgments. And, for emotive captions inspired by language use on social media, even \reqref metrics fall short.

\section{Related Work}

\paragraph{Reference-only image caption evaluation} In general, image caption generation models are evaluated by a suite of 5 reference based metrics: \bleu~\cite{papineni2002bleu} (which measures a version of precision between a candidate and the references), \rouge~\cite{lin2004rouge} (which measures a version of recall), \meteor~\cite{banerjee2005meteor} (which computes a word-level alignment), \cider~\cite{vedantam2015cider} (which combines n-gram tf-idf weighting and stemming) and \spice~\cite{anderson2016spice} (which applies a semantic parser to a set of references, and computes similarity using the predicted scene graph).\footnote{For comparison with these metrics, we use the standard COCO evaluation tools available at \url{https://github.com/tylin/coco-caption}.} \newcite{yi2020improving} give a method for re-weighting BERTScore \cite{zhang2019bertscore} specifically tuned to the image caption generation domain (we refer to their method as \bertscoretbr). 

\paragraph{Reference+image caption evaluation} Recent metrics incorporate image-text grounding features in addition to references: \tiger \cite{jiang2019tiger} uses a pretrained SCAN model \cite{lee2018stacked}, and \vilbertscore \cite{lee2020vilbertscore} uses a pretrained ViLBERT model \cite{lu2019vilbert} that is also fine-tuned on 12 downstream vision and language tasks \cite{lu202012}. Our work provides perspective on the next logical extension: instead of incorporating visual-textual interactions in addition to references, can we ignore references entirely?

\paragraph{Self-retrieval for image captioning} Prior works have proposed incorporating a \emph{self-retrieval} loss into caption generation, with the intuition that good captions should be able to uniquely identify their images with high accuracy \cite{dai2017contrastive,luo2018discriminability,liu2018show}; monitoring this type of loss
can provide insight into how distinctive the captions are according to the model itself. \clipscore is similar in spirit, but distinct for its utility as an extrinsic evaluation metric like \bleu or \cider.

\paragraph{Reference-free evaluation} In addition to the machine translation cases highlighted in the introduction, reference-free evaluations have been proposed for other generation tasks, including summarization \cite{louis2013automatically,peyrard2018objective,sun2019feasibility} and dialogue \cite{tao2018ruber,mehri2020usr}. These metrics can be supervised, relying on human judgments for quality estimation, or less-supervised, relying on pre-trained model representations. For image captioning, a version of VIFIDEL \cite{madhyastha-etal-2019-vifidel} was proposed for reference-free evaluation; however, VIFIDEL, computed based on a list of detected objects in the image from a fixed object vocabulary, generally produces less correlation with human ratings vs. reference-based metrics.

\section{\clipscorelong}

\paragraph{Model Details.} CLIP \cite{radford2learning} is a cross-modal retrieval model trained on 400M (image, caption) pairs gathered from the web. 500K search queries, consisting of common unigram/bigrams, named entities, etc., were executed on a search engine. For each query, up to 20K (image, caption) pairs were collected.

The model we use is the \texttt{ViT-B/32} version.\footnote{We expect that more powerful, larger versions of the model, if released at a later date, could perform better.} It represents images via a Vision Transformer \cite{vaswani2017attention,dosovitskiy2020image}, which forgoes convolutional filters in favor of self-attention maps computed between a 7 by 7 grid of image patches, which evenly divides a 224 by 224 pixel input image. This model has 12 transformer layers and 86M parameters. The text is similarly represented by a 12-layer transformer trained over a vocab of 49K BPE token types \cite{sennrich2015neural} (and is more fully described in \newcite{radford2019language}). Both the text and image networks output a single vector; these vectors aim to represent the content of an input caption or an image, respectively. In the case of \texttt{ViT-B/32}, these vectors are 512-D. The model's weights are trained to maximize the scaled cosine similarity of truly corresponding image/caption pairs while simultaneously minimizing the similarity of mismatched image/caption pairs using InfoNCE \cite{sohn2016improved,oord2018representation}. We hold fixed this set of weights for our experiments. %

\paragraph{Evaluating Caption Generations with CLIP.} To assess the quality of a candidate generation, we pass both the image and the candidate caption through their respective feature extractors. Then, we compute the cosine similarity of the resultant embeddings.\footnote{More sophisticated CLIP configurations, e.g., region-level/token-level correspondence models, did not achieve better performance.} We found that prefixing candidates with the prompt: ``A photo depicts" improved correlations slightly (and is our recommended/standard configuration), though ``A photo of", the recommended prompt from \newcite{radford2learning}, worked well too. Following \newcite{zhang2019bertscore}, we perform a re-scaling operation.\footnote{While the cosine similarity, in theory, can range from $[-1, 1]$ (1) we never observed a negative cosine similarity; and (2) we generally observe values ranging from roughly zero to roughly $.4$. The particular value of $w$ we advocate for, $w=2.5$, attempts to stretch the range of the score distribution to $[0,1]$. For more details and justification for our re-scaling, including a demonstration of generality across several corpora, see \autoref{appendix:rescaling_details}).} For an image with visual CLIP embedding $v$ and a candidate caption with textual CLIP embedding $c$, we set $w=2.5$ and compute \clipscore as:
\definecolor{posterR}{RGB}{0,0,0}
\definecolor{posterC}{RGB}{0,0,0}
\definecolor{posterV}{RGB}{0,0,0}
\begin{align*}
\texttt{\clipscore}(\textcolor{posterC}{\textbf{c}}, \textcolor{posterV}{\textbf{v}}) =  w * \max(cos(\textcolor{posterC}{\textbf{c}}, \textcolor{posterV}{\textbf{v}}), 0)
\end{align*}
To compute corpus-level \clipscore, we simply average
over (candidate, image) pairs.
Note that this evaluation \emph{does not depend on underlying references.} The runtime of \clipscore with the \texttt{ViT-B/32} backbone is fast: on our single consumer GPU and hard drive, roughly 4K image-candidate pairings can be processed per minute.

\paragraph{\refclipscorelong} \clipscore can additionally be extended to incorporate references, if they are available. We extract vector representations of each available reference by passing them through CLIP's text transformer; the result is the set of vector representation of all references, $R$. Then, \refclipscorelong is computed as a harmonic mean of \clipscore, and the maximal reference cosine similarity, i.e., 
\begin{align*}
& \texttt{\refclipscore}(\textcolor{posterC}{\textbf{c}}, \textcolor{posterR}{\textbf{R}}, \textcolor{posterV}{\textbf{v}}) = \\ 
& \text{H-Mean}( \texttt{\clipscore}(\textcolor{posterC}{\textbf{c}}, \textcolor{posterV}{\textbf{v}}), \max(\max_{\textcolor{posterR}{\textbf{r}} \in \textcolor{posterR}{\textbf{R}}} cos(\textcolor{posterC}{\textbf{c}}, \textcolor{posterR}{\textbf{r}}), 0))
\end{align*}

\section{Benchmark Captioning Evaluations}

\label{sec:sec_with_basic_experiments}

We first evaluate on a set of literal description corpora. Broadly, the captions in these corpora aim to identify and highlight the literal, salient objects/actions in a photographic image, presented without additional context.\footnote{See \newcite{berg2012understanding} for a statistical exploration of salience in a such a corpus.}

\subsection{Caption-level likert judgments} We first explore three corpora consisting of human likert-scale judgments at the level of individual image/caption pairs. Flickr8K-Expert \cite{hodosh2013framing} contains 17K ``expert" human judgments between 5664 images: humans graded captions on a scale of 1 to 4 (4=``caption describes the image without any errors"; 1=``caption is unrelated to the image"). Flickr8K-CF is a set of 145K binary quality judgments gathered from CrowdFlower over 48K (image, caption) pairs (1K unique images). Each pair has at least 3 binary judgments, and we take the mean proportion of ``yes" annotations as a score for each pair to compute correlations.

Composite \cite{aditya2015images} contains 12K human judgments between images from MSCOCO (2007 images), Flickr8k (997 images), and Flickr30k \cite{young2014image} (991 images). Each image originally has five references, but one of the references was selected to be rated by humans in the set (and so we remove it from the reference set when computing metrics; this differs from some prior work, see \autoref{appendix:correlation_details} for why we consider the more difficult setting). For Composite and Flickr8K judgments, we compute correlation between each metric and the human ratings using Kendall $\tau$.

\begin{table}[t]
    \centering
    \resizebox{.7\linewidth}{!}{
\begin{tabular}{lcc}
\toprule
           & $\tau_c$ \\
           \midrule
 \bleuone    & 32.3   \\
 \bleufour    & 30.8  \\
 \rouge     & 32.3 \\
 \bertscorerobertaf & 39.2 \\
 \meteor    & 41.8 \\
 \cider     & 43.9  \\
 \spice     & 44.9 \\
 \leic ($\tau_b$)* \cite{cui2018learning} & 46.6\\
 \bertscoretbr \cite{yi2020improving} & 46.7 \\
 \tiger \cite{jiang2019tiger} & 49.3 \\
 \metric{NUBIA}* \cite{kane2020nubia} & 49.5 \\
 \vilbertscore \cite{lee2020vilbertscore} & 50.1 \\
 \midrule
 \clipscorenorefs & 51.2 \\
 \refclipscore & \textbf{53.0} \\
\bottomrule
\end{tabular} }

    \caption{Flickr8K-Expert correlations with human judgment. All metrics use 4-5 ground truth references, except for \clipscore (which uses none). * indicates a result reported in prior work.}
    \label{tab:flickr8k}
\end{table}

\paragraph{Results} The results for Flickr8K-Expert are given in \autoref{tab:flickr8k}, for Flickr8K-CF are given in \autoref{tab:flickr8k_crowdflower} (in $\tau_b$, following \newcite{cui2018learning}), and for Composite are given in \autoref{tab:composite}. For the caption-level corpora we consider, \clipscore without references achieves higher correlation with human judgment compared to previously proposed metrics that rely on references. Additionally, in all cases, \refclipscore improves correlation even further. This provides strong evidence that, in terms of correlating with human judgment at the caption-level for these literal photographic image description tasks, a relatively direct application of CLIP can serve as a strong automatic evaluation metric.

\begin{table}[t]
    \centering
    
\resizebox{.55\linewidth}{!}{\begin{tabular}{lc}
\toprule
     &  $\tau_b$ \\
\midrule
 \bleufour  & 16.9  \\
 \cider     & 24.6 \\
 \meteor    & 22.2  \\
 \rouge     & 19.9  \\
 \spice     & 24.4  \\
 \bertscorerobertaf & 22.8 \\
 \leic*      & 29.5 \\
 \midrule
 \clipscorenorefs & 34.4  \\
 \refclipscore & \textbf{36.4}  \\
\bottomrule
\end{tabular}}

    \caption{Flickr8K-CF correlations with human judgment. * indicates a result reported in prior work.}
    \label{tab:flickr8k_crowdflower}
\end{table}

\subsection{Pairwise ranking on Pascal-50S}

In Pascal-50S \cite{vedantam2015cider}, raters made pairwise preference judgments between pairs of sentences. There are 4K sentence pairs total, split evenly across four categories, e.g., two human captions, two machine captions, etc. For each pair, 48 human pairwise judgments were gathered.\footnote{Instead of being presented with the image, annotators were presented only with a reference (and the two candidates to rank).} Following prior work, instead of computing correlation coefficients, we compute accuracy, i.e., we consider the caption preferred by a majority of annotators to be correct, and measure how often the evaluation metric assigns a higher score to that member of the pair. Ties are broken randomly. Due to random selection of 5 references among the 48 candidates to serve as ground-truth for the \reqref metrics, the results may differ slightly from prior work (we average over 5 random draws of references).

The results are given in \autoref{tab:pascal_50s}. Evaluation is split across four categories of caption pairs (detailed in the table caption). \clipscore and \refclipscore generally achieve high performance in all categories.

\subsection{System-level correlation for MSCOCO} \clipscore achieves high correlation with human judgments at the system-level as well: we evaluate the outputs of systems submitted to the 2015 MSCOCO Image Captioning Challenge \cite{vinyals2016show}. We have some concerns with standard evaluation setup on this corpus, mostly related to the fact that it consists of only 12 datapoints (see supplementary for more discussion). Nonetheless, following the standard procedure, we correlate \clipscore and \refclipscore with two metrics: ``the percentage of captions that are evaluated as better or equal to a human caption (M1)" and percentage of captions that pass the ``Turing Test" (M2), respectively. \clipscore achieves Spearman $\rho_{M1}/\rho_{M2}=.59/.63$ and \refclipscore achieves $\rho_{M1}/\rho_{M2}=.69/.74$ (all $p<.05$) with these system-level metrics.

\begin{table}[t]
    \centering
    \begin{tabular}{lc}
\toprule
  & $\tau_c$ \\
  \midrule
 \bleuone    & 31.3    \\
 \bleufour    & 30.6    \\
 \rouge     & 32.4    \\
 \bertscorerobertaf & 30.1 \\
 \meteor    & 38.9    \\
 \cider     & 37.7    \\
 \spice     & 40.3    \\
 \bertscoretbr* & 44.9 \\
 \tiger     & 45.4    \\
 \vilbertscore & 52.4 \\
 \midrule
 \clipscorenorefs & 53.8    \\
 \refclipscore & \textbf{55.4} \\
\bottomrule
\end{tabular}
    \caption{Composite correlations with human judgment. All metrics use between 4 and 5 ground truth references, except for \clipscore (which uses none). In contrast to some prior work, we consider a harder setting, and remove the candidate from the reference set (see \autoref{appendix:correlation_details} for details; for comparison purposes, \refclipscore achieves $\tau_c=60.0$ in the easier setting). * indicates a result reported in prior work.}
    \label{tab:composite}
\end{table}

\begin{table}[t]
    \centering
    \resizebox{\linewidth}{!}{
\begin{tabular}{lcccc|c}
\toprule
                &   HC &   HI &   HM &   MM &   Mean \\
\midrule
 length               & 51.7 & 52.3 & 63.6 & 49.6 &   54.3 \\
 \bleufour              & 60.4 & 90.6 & 84.9 & 54.7 &   72.6 \\
 \spice               & 63.6 & 96.3 & 86.7 & 68.3 &   78.7 \\
 \meteor               & 63.8 & 97.7 & 93.7 & 65.4 &   80.1 \\
 \rouge                & 63.7 & 95.3 & 92.3 & 61.2 &   78.1 \\
 \cider                & 65.1 & 98.1 & 90.5 & 64.8 &   79.6 \\
 \bertscorerobertaf                 & 65.4 & 96.2 & 93.3 & 61.4 &   79.1 \\
 \midrule
 \tiger* & 56.0 & \textbf{99.8} & 92.8 & 74.2 & 80.7 \\
 \vilbertscore* & 49.9 & 99.6 & 93.1 & \textbf{75.8} & 79.6 \\
 \bertscoretbr* & \textbf{65.4} & 98.1 & \textbf{96.4} & 60.3 & 80.1 \\
 \midrule
 \clipscorenorefs            & 56.5 & 99.3 & \textbf{96.4} & 70.4 &   80.7 \\
 \refclipscore         & 64.5 & 99.6 & 95.4 & 72.8 &   \textbf{83.1} \\
\bottomrule
\end{tabular}}

    \caption{Pascal50S accuracy results (5 references). HC = two human correct captions; HI = both captions are human written, but one is wrong; HM = both captions are for the image, but one is written by a human, one by an algorithm; MM = both captions are for the image, and both are written by an algorithm. * indicates a result reported in prior work: the comparability of our results to *-rows is subject to the (arbitrary) sample of references. We average our results over 5 random samples (but \clipscore doesn't change because it doesn't use references). }
    \label{tab:pascal_50s}
\end{table}

\begin{table}
    \centering
    
\begin{tabular}{lcc}
\toprule
& 1-ref & 4-ref \\
\midrule
 length                                     & 50.2 & 50.2 \\
 \bleufour                                   & 66.5 & 82.6 \\
 \meteor                                    & 78.8 & 85.4 \\
 \rouge                                     & 71.7 & 79.3 \\
 \cider                                     & 82.5 & 90.6 \\
 \spice                                     & 75.5 & 86.1 \\
 \bertscorerobertaf                 & 88.6 & 92.1 \\
 \midrule
 \clipscore (no refs)                                 & 87.2 & 87.2 \\
 \refclipscore                              & \textbf{91.0} & \textbf{92.6} \\
\bottomrule
\end{tabular}
    \caption{Accuracy of evaluation metrics in the pairwise FOIL hallucination detection setting. All \reqref metrics are given access to either one or four references.}
    \label{tab:FOIL}
\end{table}

\subsection{Sensitivity of \clipscore to hallucination} Prior work has demonstrated that, for many literal description tasks, humans often prefer \emph{correctness} in captions over specificity \cite{rohrbach2018object,rohrbach2017movie}.\footnote{This is not always the case: \newcite{macleod2017understanding} show there is a range of opinion among a sample of low vision and blind users of social media.} Thus, understanding if and how evaluation metrics handle image captions that contain incorrect ``hallucinations," e.g., references to objects that are not depicted, is important. We use a sample of image captions from the FOIL dataset, constructed by \newcite{shekhar2017foil}, to test how sensitive \clipscore is to detecting potentially subtle inaccurate details in descriptions. This corpus consists of modified reference captions from MSCOCO that have a single noun-phrase adversarially swapped out to make the FOIL caption incorrect, e.g., switching ``motorcycle" for ``bicycle".

To adapt the corpus to our setting, for each of the 32K test images, we sample a (FOIL, true) pair, and compute the accuracy of each evaluation metric in their capacity to assign a higher score to the true candidate versus the FOIL. To compute \reqref metrics, we give access to the MSCOCO reference captions for the image (excluding the the true candidate being assessed against the FOIL). While the paired setting we consider isn't identical, \newcite{shekhar2017foil} estimate roughly 92\% human agreement on the unpaired version of the task, relative to a 50/50 random guessing baseline.

\autoref{tab:FOIL} contains the results. In this setting, having access to more annotation is quite helpful for reference based metrics, e.g., the accuracy of \spice and \bleufour increase by over ten points when shifting from one to four references. But in the reference-limited setting, \clipscore, without \emph{any} reference outperforms all metrics except for \bertscorerobertaf. And, \refclipscore works best in all cases.

Overall, we corroborate \newcite{rohrbach2018object}'s finding that ``object hallucination can not be always predicted
based on the traditional sentence metrics" using a corpus derived from \newcite{shekhar2017foil}, particularly in the case where there are few references available. However, \clipscore and \refclipscore offer a performance improvement in the pairwise setting. 

\subsection{Sensitivity of \clipscore to memorization}

\label{sec:sec_with_memorization_test}

One concern with model-based scoring methods is memorization, i.e., if a model's weights are pre-trained using a large corpus, there's a risk that data used at evaluation time have already been seen at pretraining time. While \newcite{radford2learning} conduct a train-test overlap analysis and find that CLIP is unlikely to succeed because of memorization, we nonetheless conduct an experiment with images CLIP has never seen before.

The authors of this work created a set of 250 images that have never been posted to the Internet by aggregating personal photographs. The set contains a variety of Flickr-like situations, e.g., nature scenes, animals, city streets, objects, etc. For each image, we collect two automatically generated captions: one from a commercial API, Microsoft Azure Cognitive Services (v 3.1)\footnote{\url{https://azure.microsoft.com/en-us/services/cognitive-services/}} and one from \newcite{luo2018discriminability}'s pretrained model, which is trained to maximize \cider score with a self-critical baseline.\footnote{We use the ResNet101 pretrained version, which achieves 1.05 \cider and 0.19 \spice on the COCO validation set.} Then, for each image, three authors of this work independently selected which caption described the image content more accurately.  Relative to a 50\% random baseline (and a 72\% length baseline of selecting the \emph{shorter} caption) \clipscore correctly recovers majority human preference in 86\% of cases. Human agreement for this corpus is 93\%.\footnote{Raters preferred the Microsoft captions to the ResNet101 model 81\% of the time.}

While this setup cannot definitively refute the notion that CLIP works well because it has memorized images, we hope the results here contribute to the evolving discussion about the nature of generalization for web-scale pretrained models.

\subsection{Which metrics should I report?}
\label{sec:sec_with_correlation}

\begin{figure}
    \begin{subfigure}{.49\linewidth}
    \centering
     \includegraphics[width=.98\linewidth]{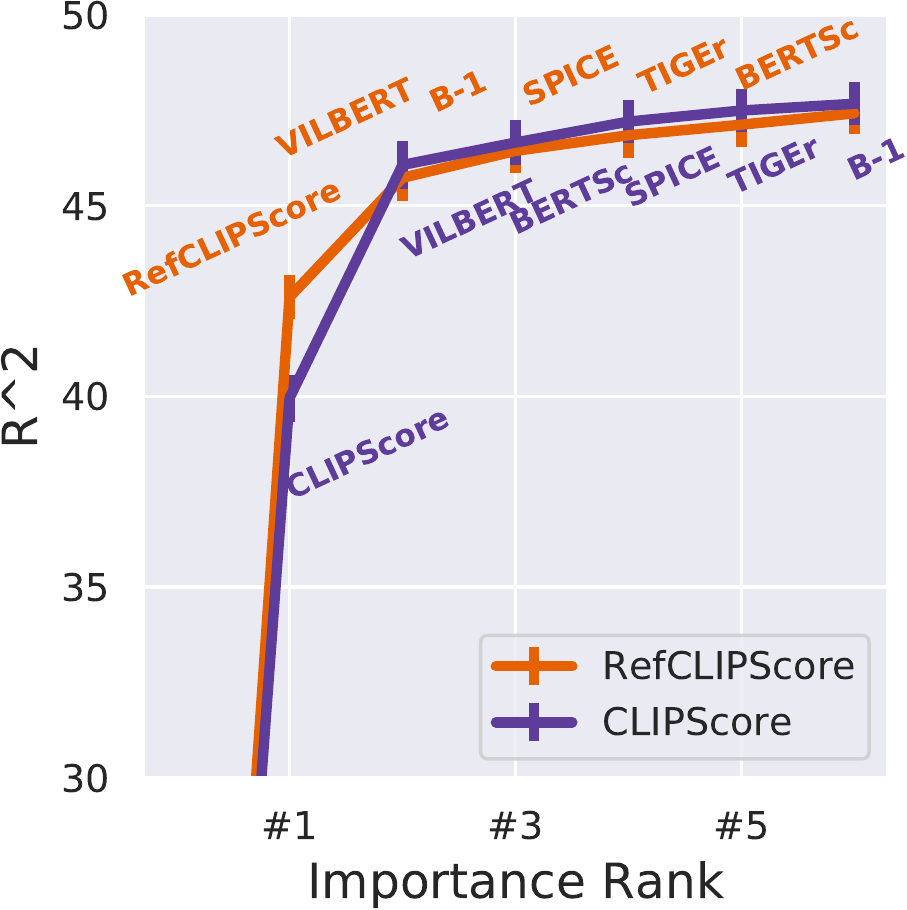}
     \caption{Composite}
    \end{subfigure} \hfill %
    \begin{subfigure}{.49\linewidth}
     \centering
     \includegraphics[width=.98\linewidth]{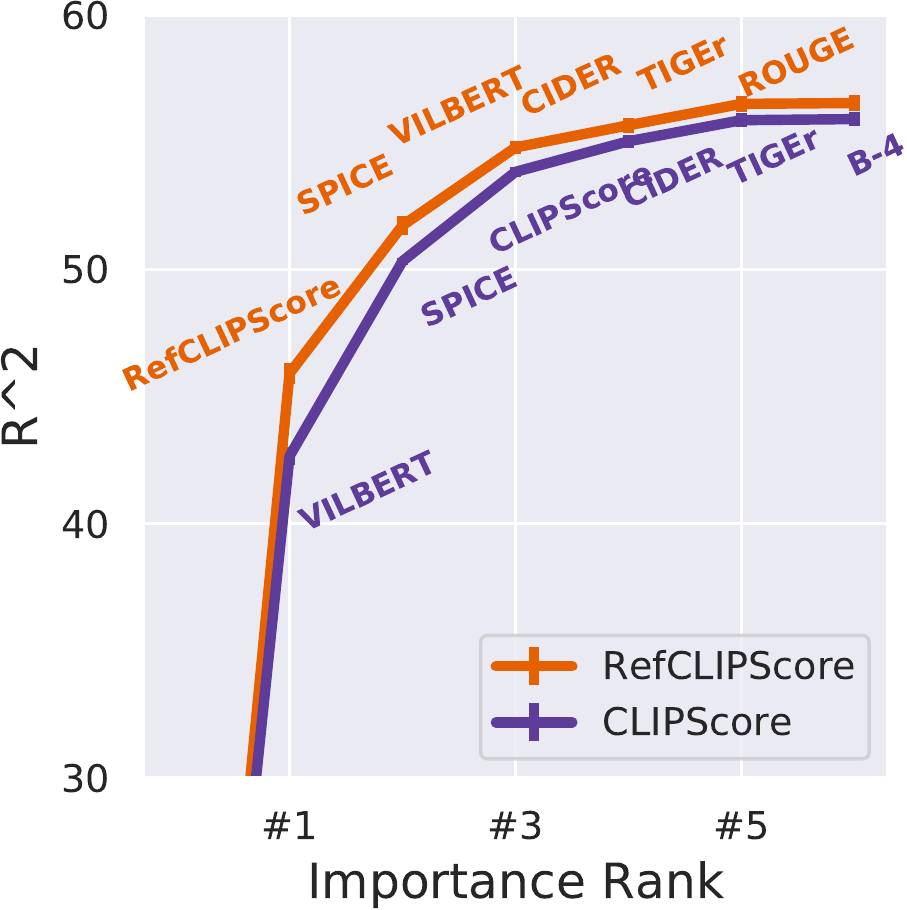}
     \caption{Flickr8k-Expert}
    \end{subfigure}
    \caption{$R^2$ for the forward-selection regression of metrics on human Likert ratings for two corpora. Foward-selection tends to identify both {\textbf {\color{cbpurple} \clipscore}} and \textbf{{\color{cborange} \refclipscore}} early-on: other informative and complementary metrics include \vilbertscore and \spice.}
    \label{fig:forward_selection}
\end{figure}

Most caption generation works report multiple metrics, each of which (presumably) correlates with human judgment to different degrees. But it's not always clear if individual metrics capture distinct or redundant dimensions of human judgment. For example, while \clipscore and \vilbertscore both produce high correlations, are they redundant or complementary?

We seek a (minimal) set of metrics that explains the most variance in human judgment. To find this set, we undertake a forward selection on a set of ten candidate metrics comprising six widely-reported metrics,\footnote{\bleuone, \bleufour, \meteor, \cider, \rouge, \spice} and four newer metrics, 
\bertscorerobertaf, \tiger, \vilbertscore, and \clipscore (we also include experiments starting with \refclipscore instead of \clipscore, too).
Starting from an empty set, we perform an iterative greedy selection by picking the most informative additional metric to add.\footnote{Our criteria is how much additional $R^2$ correlation with human judgment a metric adds according to a linear regression. We use \texttt{sklearn} \cite{scikit-learn}'s forward selection, which applies 5-fold cross-validation at each step.} To estimate variance, we repeat the forward-selection process 10 times with bootstrap re-sampled versions of the corpus.

\autoref{fig:forward_selection} shows the information gain that results from running this experiment on the Composite and Flickr8K-Expert corpora; we also show which metric is most commonly selected at each iteration (earlier = more information gain). For Composite, \clipscore (or \refclipscore) is always selected first, followed by \vilbertscore, and then (most commonly) \bertscorerobertaf. For Flickr8k-Expert, the top three choices are always \clipscore (or \refclipscore), \vilbertscore, and \spice. While \clipscore and \vilbertscore tend to be the most informative metrics, (1) while they are correlated, they are not purely redundant; and (2) image-unaware, \reqref metrics like \spice can still be useful.

In summary, these results suggest that evaluation metrics like \clipscore, which take into account visual content, indeed capture axes of human judgment not currently covered by text-only reference-based metrics. \emph{For the literal image description evaluation settings we consider, a reasonable mix of metrics to report is at least one image-aware metric (e.g., \clipscore) plus a strong reference-only metric (e.g., \spice).}

\section{Case Studies Using \clipscorelong}

\label{sec:sec_with_divergent_domains}

Our results thus far have demonstrated that CLIP encodes information useful for evaluating literal image description tasks. But, \reqref metrics may \emph{a priori} seem more adaptable versus \clipscore. Does \clipscore correlate with human judgment beyond cases like MSCOCO and Flickr8K?

\begin{figure}
    \centering
    \includegraphics[width=.99\linewidth]{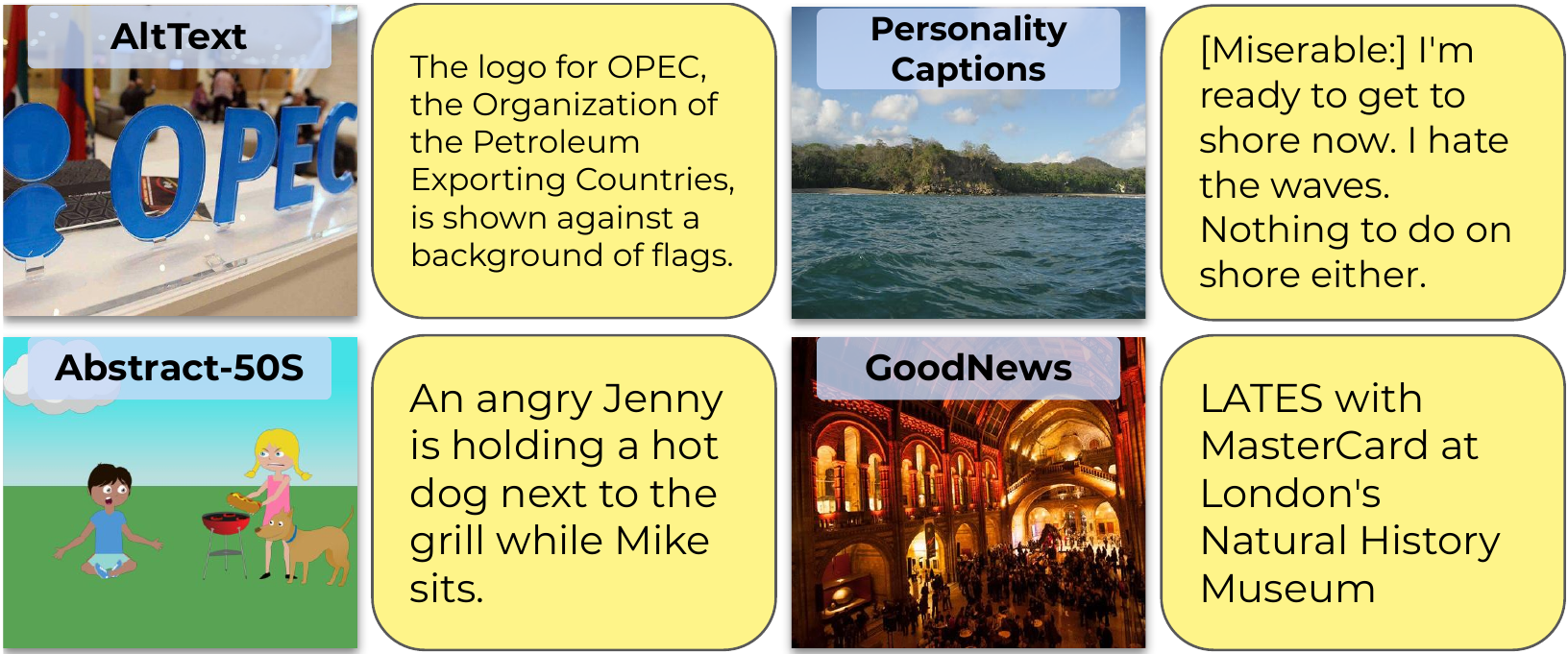}
    \caption{Instances from our four case-study corpora.}
    \label{fig:new_domain_examples}
\end{figure}

To address this question, we consider four case studies, exploring the correlation between \clipscore and human judgment across ``divergent" image description datasets. These corpora qualitatively differ from the more popular domains explored in \S \ref{sec:sec_with_basic_experiments}, either because the images are not ``everyday" images from Flickr, or because the captions are not literal description (\autoref{fig:new_domain_examples} illustrates).

\subsection{Alt-Text ratings from Twitter}

When uploading an image alongside a tweet, users of Twitter have the option of providing alternative text: while few use this feature (\newcite{gleason2019s} find that fewer than .1\% of image tweets have alt-text), its broader adoption might someday make social media more accessible for low vision and blind users. We measure \clipscore's capacity to reconstruct a set of 2.8K human judgments of alt-text quality. This corpus was collected and rated by the authors of \newcite{gleason2019s,gleason2020twitter}. Each alt-text was rated on a scale of 0 to 3 in terms of its probable utility as an alt-text. While the human-raters raters themselves are sighted thus cannot directly assess the utility of a given alt-text to a low vision or blind user, they are experts in designing and evaluating alt-text systems. Tweets were sampled from a mix of the Twitter FireHose API, and the timelines of low vision and blind users of the site. The images, qualitatively, are a broader mix of web content in comparison to Flickr-like domains, e.g., screenshots, memes, etc. Alt-text candidates are a mix of user-uploaded and machine-generated. The corpus contains no references, but for the purposes of comparison to \reqref metrics, we (programmatically) treat any textual context of the tweet as a reference.

\clipscore achieves 48.4 $\tau_c$ correlation with the human judgements. In contrast, likely due to the unreliability of Tweet texts as viable alt-texts, \reqref methods struggle: the best performing purely-reference based metric, \bertscorerobertaf (which achieves 15 $\tau_c$) under-performs relative to length baseline (which achieves 25 $\tau_c$). While gathering high-quality, contextual reference alt-texts is a promising avenue for future work,\footnote{See \newcite{stangl2020person}, who conducted user-studies across six domains.} \clipscore offers a promising evaluation metric candidate in this domain.

\subsection{Abstract-50S}

We assess \clipscore's capacity to generalize to abstract, non-photographic clip-art images using Abstract-50S \cite{vedantam2015cider}. This dataset pairs clip-art images (originally constructed by \newcite{zitnick2013bringing}) with 48 human-written reference captions. These images depict two cartoon characters, Mike and Jenny, in various outdoor situations, e.g., playing sports, having a picnic, etc. For 400 human-written candidate caption pairs (200 pairs are from the same image, 200 are from different images), human judgments were collected: annotators were instructed to choose which of the paired captions were more similar to each reference caption, so 48 judgments were collected for each candidate pair (for a total of 19200).

We compare \clipscore to several \reqref metrics when given access to a random sample of five reference captions. Following our procedure for Pascal-50S, we randomly re-sample 5 times, and report average pairwise accuracy. Two baselines (BL) both achieve 53: length-only (i.e., saying the longer caption is better); and randomly shuffling images as input to \clipscore (so that it cannot rely on meaningful visual-textual interactions).

\begin{center}
\resizebox{.98\linewidth}{!}{
\begin{tabular}{cccccc}
  BL & \bleu & \cider & \meteor & \bertscore & \clipscore (no refs)\\
  \midrule
  53 & 71 & 79 & 79 & 73 & 68 \\
\end{tabular}
} \end{center}

Overall, while \clipscore underperforms relative to the \reqref metrics, it outperforms the baselines by a wide margin. This result suggests that \clipscore is capable of reasoning about visual-textual interactions, even in non-photographic images.

\subsection{Personality Captions} 

Inspired by language use on social media, \newcite{shuster2019engaging} collected image captions by prompting annotators with a ``personality" (e.g., dramatic, sympathetic, sad, etc.) and asking them to ``write a comment in the context of [a] given personality trait... about an image that someone else would find engaging." %
To evaluate their models, the authors collected pairwise human judgments, where evaluators were instructed to ``to pick which comment is the most engaging". We assess \clipscore in two capacities: (1) does it prefer literal descriptions, or the less-literal, more engaging, personality captions?; and (2) if it is given two personality captions, can it predict which humans judge to be more engaging?

For (1): Over a set of 2.4K ``traditional" vs. personality captions pairwise ratings, humans rate the personality captions to be more engaging 65\% of the time, whereas \clipscore prefers the traditional 80\% of the time.\footnote{Preliminary prompt-engineering experiments (e.g., ``when I look at this photo, I feel [PERSONALITY] and think [CAPTION]") could not overcome this.} Our takeaway: when given a direct description and a more engaging, non-literal caption, \clipscore will generally prefer the literal.

For (2): \clipscore performs slightly better than random, e.g., 57\% over 2.5K human pairwise judgments comparing two neural generator models: TransResNet (ResNeXt-IG-3.5B) vs. TransResNet (ResNet-152) (see \newcite{shuster2019engaging} Table 7, Row 5), but no better than a length-only baseline (also 57\%). Notably, even \reqref metrics fail to provide correlation with pairwise human judgment of engagingness on this corpus: e.g., \bleufour, \cider, and \spice agree with human judgment 52\%/53\%/51\% when provided with one personality-primed reference. Our takeaway: when given two engaging, non-literal descriptions, both \clipscore and traditional \reqref metrics fail to predict which humans will judge to be more engaging.

\subsection{News image captioning}
\label{sec:with_news_captioning}

\newcite{biten2019good} consider caption generation for images from New York Times articles; their task differs from MSCOCO because 1) 95\% of captions contain at least one named entity, e.g., a politician, celebrity, or place; and 2) captions generally ``do not describe scene objects, but rather offer a contextualized interpretation of the scene."
They collected 2.1K pairwise human judgments over 106 images that compare the performance of two news image captioning models. For each image, 20 annotators were instructed to pick which of two model generations was closer to the ground-truth caption (they were also presented with the image itself). We compare metrics in terms of their accuracy in matching human judgment between the two candidates.

\Reqref metrics dominate: \meteor and \bleufour achieve the highest accuracies of 93 and 91 respectively, whereas \clipscore achieves only slightly above random at 65. Qualitatively, \clipscore succeeds when there are visually-verifiable content, e.g., matching black-and-white photos to older dates (e.g., picking 1933 vs. 1977, in one case), and matching particularly iconic celebrities (e.g., it confidently identifies Muhammad Ali boxing).\footnote{\newcite{luo2021newsclippings}'s recent experiments quantitatively demonstrate that CLIP is capable of reasoning about real-world entities within news images.} But, its most common failure case are captions that may simply be unverifiable given only the image content. For example: \clipscore selects ``The dining room at Elle Decor" for an image of a room, but annotators preferred a caption that mentioned ``the Junior League of New York;" the ground truth caption reveals why the image was pictured in the first place: ``A Manhattan home on a May 7 tour by the Junior League of New York."

Overall, we do not advocate for reference-free evaluation in this case, especially because our results suggest that (at least for this particular set of annotations) \reqref n-gram overlap metrics achieve high correlation with human judgment.

\section{Conclusion}

For literal image description tasks, \clipscorelong achieves high correlation with human judgments of caption quality \emph{without} references when used in an off-the-shelf fashion. Additional experiments in divergent domains suggest that CLIP can also reason about non-photographic clip-art, and serves as a reasonable option for reference-free evaluation in the alt-text case. Promising future work includes exploring 1) \clipscore as a reinforcement learning reward for literal caption generators; and 2) whether a small amount of labelled human rating data could help \clipscore adapt to domains where it struggles, e.g., engagingness prediction. We hope our work can contribute to the ongoing discussion about the role of pretrained models in generation evaluation.

Reference-free evaluation runs some risks. Much like BERTScore, model-based metrics like \clipscore reflect the biases of the pre-training data. While we believe that using \clipscore as an offline evaluation metric for literal caption quality accords with the recommendations of CLIP's model card\footnote{\url{https://github.com/openai/CLIP/blob/main/model-card.md}} \cite{mitchell2019model}, \newcite{agarwal2021evaluating}'s study demonstrates that CLIP can make disproportionate incorrect classifications of people, e.g., ``male images were misclassified into classes related to crime.'' Exploring potential social biases of candidate generations (as in, e.g., \newcite{hendricks2018women}) remains paramount, particularly if a system is to be deployed.

\paragraph{Contemporaneous work} While this work was under submission, two alternate reference-free evaluation metrics for image caption generation were introduced: FAIEr \cite{wang2021faier} (based on a pretrained object detector, and fine-tuned on MSCOCO) and UMIC \cite{lee2021umic}  (based on UNITER \cite{chen2020uniter}). UMIC, in particular, produces similar correlations with human judgment on the literal image description tasks (\S\ref{sec:sec_with_basic_experiments}) compared to \clipscore, but with the complementary approach of fine-tuning on synthetic negative captions. Future work would be well-suited to explore if the textual data augmentations proposed by \newcite{lee2021umic} (1) result in a metric that complements or overlaps with the non-finetuned \clipscore (\S\ref{sec:sec_with_correlation}); and (2) could be extended beyond cases of literal description (\S\ref{sec:sec_with_divergent_domains}).

\section*{Acknowledgements}
This research is supported in part by DARPA MCS program through NIWC Pacific (N66001-19-2-4031), DARPA SemaFor program, and the Allen Institute for AI. We additionally thank Ximing Lu, Swabha Swayamdipta, Youngjae Yu, and the anonymous EMNLP reviewers for the helpful comments, thoughts, and discussions. Finally, we thank Jin-Hwa Kim, who in March 2022, helped discover a now fixed discrepancy for the Pascal-50S results, see Appendix~\ref{app:sec_with_composite_details}.
\putbib[refs]
\end{bibunit}

\clearpage

\begin{bibunit}[acl_natbib]
\appendix
\section{Evaluation and Replication Details}

\label{appendix:correlation_details}

\newcite{anderson2016spice} introduced a set of corpora, metrics, and experimental settings for comparing image caption generation evaluation metrics. Perhaps unwittingly, their introduced protocols have become the accepted standard for evaluation of new caption generation metrics. However, seemingly innocuous preprocessing+reporting choices can significantly impact correlations with human judgment on these corpora. In what follows, we detail our replication efforts. Our goal was to make the experimental comparisons involving \clipscorelong reported in the main paper as fair as possible. We hope it can be useful for researchers reporting metrics on this setup going forward.

\subsection*{Flickr8K details} \label{sec:sec_with_evaluation_details}
We contacted the authors of some prior work, and did our best to re-create their evaluation settings. We uncovered two types of discrepancies when reporting on this corpus. The first discrepancy is that prior work has mixed evaluating rank correlations with kendall-C and kendall-B. These metrics handle ties differently, and ties are frequent because human Likert judgements are discretized.
The second discrepancy is the method of aggregation of human ratings. Three human ratings were gathered for 5664 (image, candidate) pairs. The majority of prior works flatten all human judgments to a single list, and report rank correlation over 5664 * 3 = 16992 instances (method A). However, another (possibly more defensible) evaluation choice is to average human ratings for each pair, and report rank correlation instead over 5664 instances (method B). The choice of aggregation method has a significant impact on correlations. For example, when we used aggregation method A and $\tau_c$ for \spice, we can exactly replicate the correlation, 44.9, originally reported in \cite{anderson2016spice}. But, if we use $\tau_c$ and instead use aggregation method B, the correlation increases to 52.9: this inflation occurs with other metrics, too.

For our results, we do our best to report all results for the most common setting: using $\tau_c$ correlation, and using aggregation method A. Thus, the results we report may differ slightly than the results reported in prior work.

\subsection*{Composite details}
\label{app:sec_with_composite_details}
For this corpus too, prior work has mixed evaluating with kendall-C and kendall-B correlations, which can have an impact, e.g., for \cider in our setting, switching from $\tau_b$ to $\tau_c$ results in an increase from 35 to 38 rank correlation. But perhaps the most impactful decision for this corpus relates to the references: each image originally has (roughly) five references. But when gathering human judgments, one of the candidate captions that was rated by humans was sampled from the references. For Flickr8k, \newcite{anderson2016spice} ``exclude 158 correct image-caption pairs where the candidate caption appears in the reference set;" this curation choice has become standard for Flickr8k. But for Composite, it's not clear if they repeated this curation choice, or not. And because of this ambiguity, it's not obvious which standard each prior work followed, either. For fair comparison, in an effort to reconstruct \newcite{anderson2016spice}, we tried both ways: removing the ground truth candidate reference, and not.

\begin{table}
    \centering
    \resizebox{\linewidth}{!}{

\begin{tabular}{lccccc}
\toprule
  & Original & $\tau_b$ no GT & $\tau_b$ w/ GT & $\tau_c$ no GT & $\tau_c$ w/ GT \\
  \midrule
 \bleuone   & 26 & 29 & 45 & 31 & 49 \\
 \bleufour  & 18 & 31 & 46 & 31 & 50 \\
 \rouge     & 28 & 30 & 48 & 32 & 49 \\
 \meteor    & 35 & 36 & 49 & 39 & 50 \\
 \cider     & 36 & 35 & 48 & 38 & 52 \\
 \spice     & 39 & 39 & 51 & 40 & 53 \\
\bottomrule
\end{tabular} }

    \caption{Attempts at replicating \newcite{anderson2016spice}'s results on the composite corpus.}
    \label{tab:replicate_composite}
\end{table}

Our efforts to replicate the exact values of \newcite{anderson2016spice} are in Table~\ref{tab:replicate_composite}. We suspect the discrepancy in \bleufour likely results from a smoothing issue related to the application of \bleufour to individual captions vs. the whole corpus (as mentioned in \newcite{kane2020nubia}). Based on these replication efforts, it's likely that the original evaluations for this corpus were computed using $\tau_c$ with GT references removed. We agree that the fairest analysis on this corpus should not include a reference that is also a candidate. And while we didn't go through all prior works and recompute their metrics with this change, we did compute \vilbertscore in this setting, because it was, before \clipscorelong, the state-of-the-art for this corpus. If it's helpful for future reporting: in the setting where all references (including the GT reference) are used, \refclipscore gets $\tau_c = 60.0$.

\subsection*{MSCOCO system-level details}
\label{sec:sec_with_mscoco_system_details}
The MSCOCO 2015 image captioning challenge is a standard corpus for evaluation the system-level correlation between new evaluation metrics and human judgments on the MSCOCO test set. To our knowledge, this evaluation was first conducted by \newcite{anderson2016spice} using a random sample of 1K test set submissions from 15 teams. But because the test set predictions are not public, more recent work (e.g., \newcite{cui2018learning,zhang2019bertscore}) has evaluated using dev set predictions from systems, and assuming dev set results correlate with test set results (12 teams submitted dev predictions). However, there are some potential problems with this setup:

\begin{enumerate}[leftmargin=*,topsep=0pt,itemsep=-1ex,partopsep=1ex,parsep=1ex]
    \item There's reason to believe that some teams give dev set predictions with different models vs. test set predictions. For example, the dev set predictions are identical between the two submissions: \texttt{m-RNN} and \texttt{m-RNN (Baidu/ UCLA)}, but the test set predictions differ (and achieve significantly different scores).
    \item Correlations are reported over 12 (or possibly only 11, given the duplicate predictions) systems. But spearman/pearson correlation over only 12 observations is unfortunately simple to (accidentally) ``game" due to the low statistical power of the comparison (see \newcite{card2020little} for an overview of statistical power in NLP). Consider a (nonsense) evaluation metric that assigns a random uniform $[0, 1)$ ``score" to systems without examining outputs, and consider applying this metric, e.g., $N=10$ times to the 12 systems and taking the best performing run as the final metric (simulating either a single researcher developing a new evaluation metric and/or the community's collective trials). We ran a simulation of this process 1000 times: the average spearman/pearson correlation between human judgments and our bogus metric were $r/\rho=.91$, due to repeated evaluation and low sample size.
\end{enumerate}

\noindent Thus, while the intent of this evaluation is understandable, and it may be possible to garner some insight if relatively few evaluations are conducted, this specific setup as a fine-grained comparison between new evaluation metrics for caption generation has likely outlived its utility.

\subsection*{Pascal-50S Setup Erratum}

\begin{table}[t]
    \centering
    \resizebox{\linewidth}{!}{
\begin{tabular}{lcccc|c}
\toprule
      & HC   & HI   & HM   & MM & Mean \\
      \midrule
 length &  \textbf{65.4} & 52.4 & 63.0   & 42.3 &   55.8 \\
 \bleufour  & 52.5 & 90.4 & 84.9 & 55.3 &   70.8 \\
 \spice & 56.9 & 96.3 & 87.1 & 66.4 &   76.7 \\
 \meteor & 59.0   & 97.7 & 93.9 & 62.0   &   78.2 \\
 \rouge & 55.0   & 95.3 & 93.1 & 58.7 &   75.5 \\
 \cider & 53.7 & 98.1 & 90.8 & 63.7 &   76.6 \\
 \bertscorerobertaf & 54.4 & 96.1 & 94.3 & 56.4 &   75.3 \\
 \midrule
 \clipscorenorefs & 60.3 & 99.4 & \textbf{97.9} & 77.3 & \textbf{83.7} \\
 \refclipscore & 57.9 & \textbf{99.5} & 96.1 & \textbf{80.8} &   83.6 \\
\bottomrule
\end{tabular}}

    \caption{Pascal50S-11-judgment accuracy results (5 references, non-standard 11 human judgment version). HC = two human correct captions; HI = both captions are human written, but one is wrong; HM = both captions are for the image, but one is written by a human, one by an algorithm; MM = both captions are for the image, and both are written by an algorithm. We average our results over 5 random samples (but \clipscore doesn't change because it doesn't use references).}
    \label{tab:pascal_50s_old}
\end{table}

In March 2022, Jin-Hwa Kim reported some \href{https://github.com/jmhessel/clipscore/issues/4}{small discrepancies} in a replication effort for the Pascal-50S corpus. Upon further investigation, it was discovered that the original version of this work was using a different set of human judgments than the usual setup. In particular, the \href{http://vrama91.github.io/cider/}{Pascal-50S corpus} contains two types of human judgments: 11 human judgments per pair (located in a file named \texttt{pair\_pascal.mat}); and 48 human judgments per pair (located in a file named \texttt{consensus\_pascal.mat}). The 48 judgments are intended to be used, and the results in the main paper have been updated accordingly. For reproducability sake, in case future work utilizes the 11 judgments, we have included those results in Table~\ref{tab:pascal_50s_old}.

\section{Rescaling \clipscorelong}
\label{appendix:rescaling_details}

\begin{figure}
    \centering
    \begin{subfigure}{.48\linewidth}
    \centering
     \includegraphics[width=.9\linewidth]{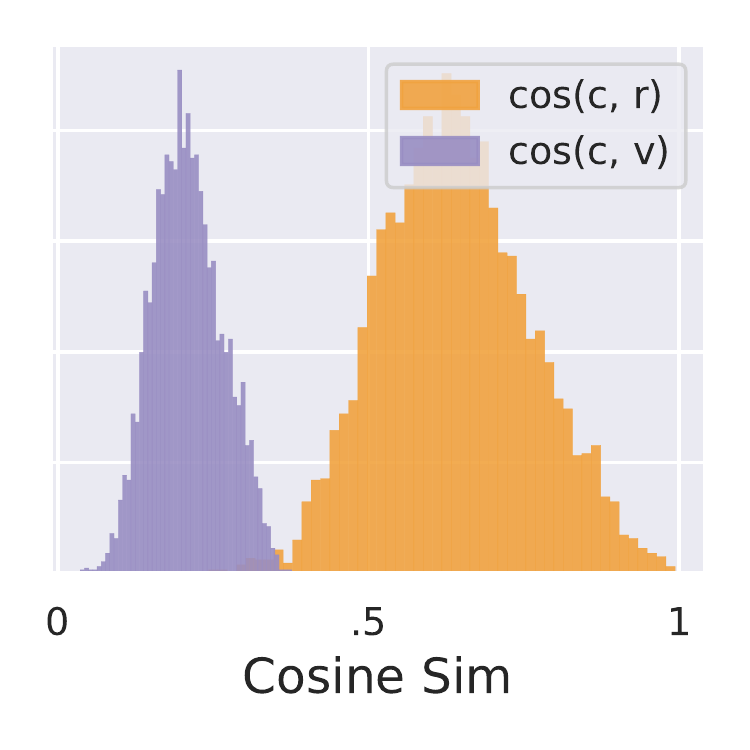}
     \caption{Flickr8K}
    \end{subfigure}%
    \begin{subfigure}{.48\linewidth}
     \centering
     \includegraphics[width=.9\linewidth]{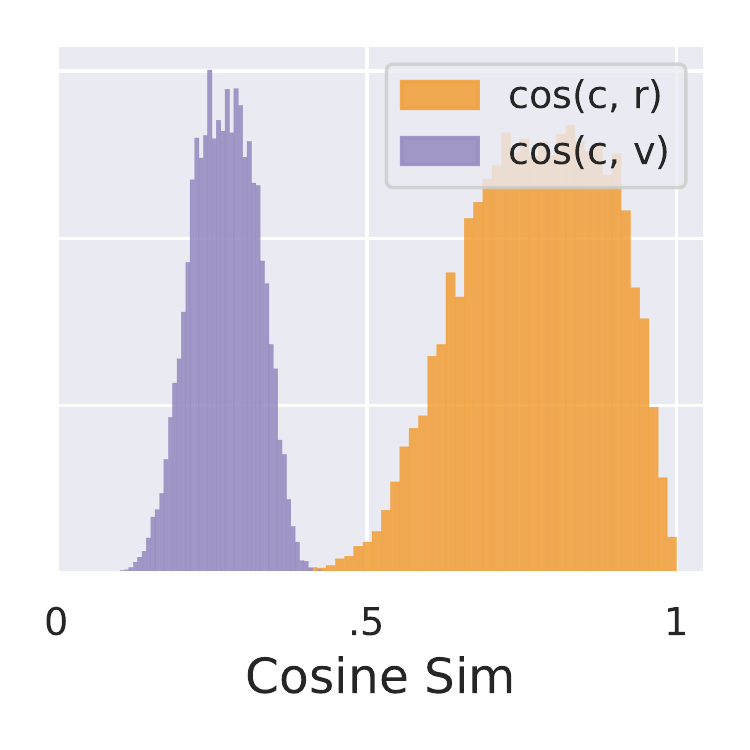}
     \caption{Composite}
    \end{subfigure}
    \centering
    \begin{subfigure}{.48\linewidth}
    \centering
     \includegraphics[width=.9\linewidth]{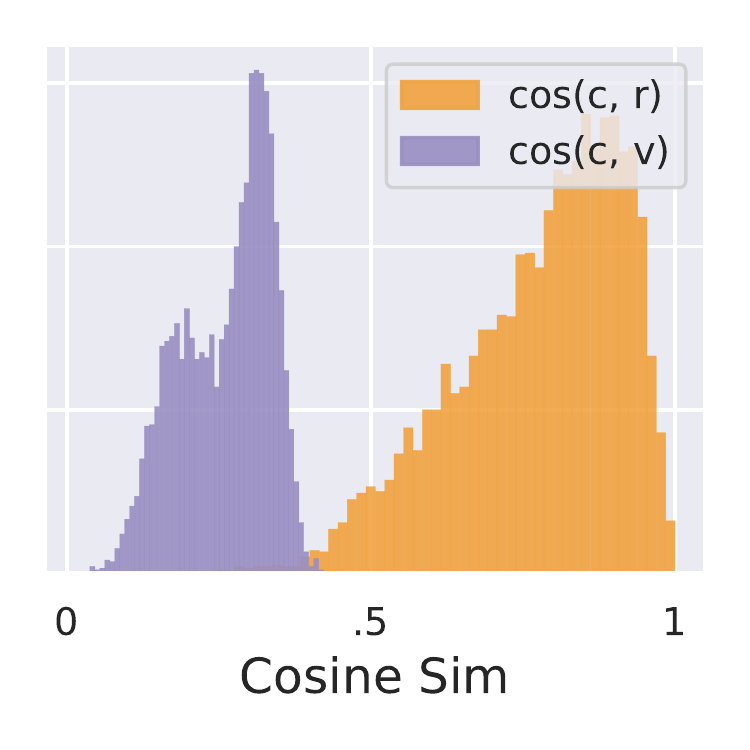}
     \caption{Pascal50S}
    \end{subfigure} %
    \begin{subfigure}{.48\linewidth}
     \centering
     \includegraphics[width=.9\linewidth]{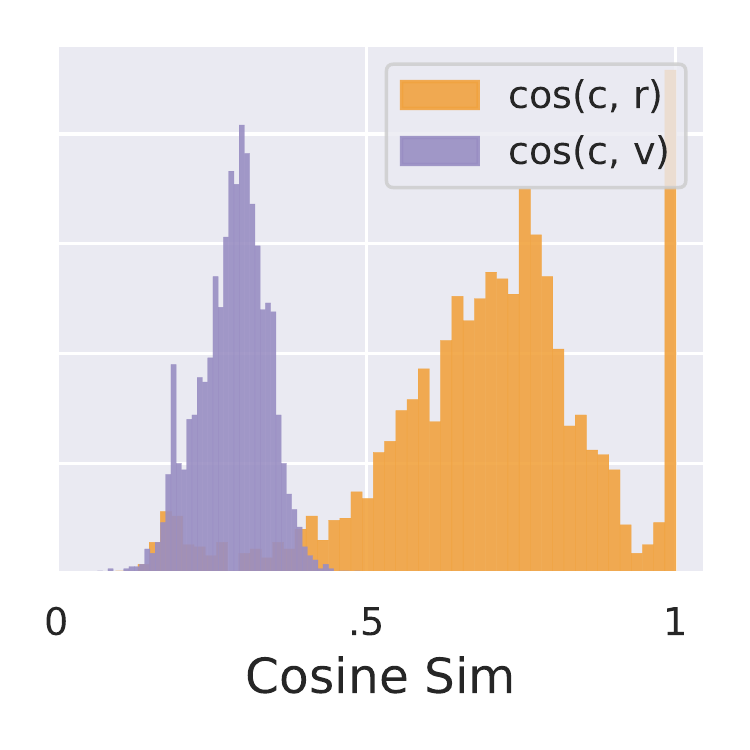}
     \caption{Twitter AltText}
    \end{subfigure}
    \caption{Distributions of raw cosine similarities between {\color{cborange} \underline{\textbf{c}}andidate and \underline{\textbf{r}}eferences} and {\color{cbpurple}\underline{\textbf{c}}andidate and \underline{\textbf{v}}isual content} from CLIP \texttt{ViT-B/32}.}
    \label{fig:cos_dists}
\end{figure}

For readability purposes, as in \newcite{zhang2019bertscore}, we sought to re-scale the raw cosine similarities computed by CLIP \texttt{ViT-B/32}.  While such a monotonic rescaling operation doesn't affect ranking results, for reporting purposes, it can be easier to compare raw values if they are on a scale more closely-aligned with other evaluation metrics (e.g., from roughly zero to roughly one). Figure~\ref{fig:cos_dists} shows the raw candidate-reference and candidate-image cosine similarities for four corpora. (Many ``reference"-candidate similarities for the Twitter corpus are 1.0 because users frequently use the text of their tweet as the AltText.) Across all of these cases, we never observed a negative negative cosine similarity. But, to be safe, we take a maximum between the cosine similarity and zero because the harmonic mean used to compute \refclipscorelong would be undefined for negative values. Multiplying by 2.5 has the effect of ``stretching" the \clipscorelong distribution to more uniformly span between zero and one, though, \clipscorelong can be greater than 1. Furthermore, when computing \refclipscorelong, we maintain this weighting, because it has the effect of mapping the visual-textual cosine similarity distribution to more closely match the reference-candidate distribution: this provides a roughly equal importance weighting between the image-candidate and reference-candidate similarity factors.

We note that the exact parameters of our rescaling method only apply to CLIP \texttt{ViT-B/32}. If future, bigger models are released, e.g., the presently unreleased \texttt{ViT-L/14} CLIP variant, they could exhibit a different cosine similarity distribution.

\putbib[refs]
\end{bibunit}


\begin{thebibliography}{69}
\expandafter\ifx\csname natexlab\endcsname\relax\def\natexlab#1{#1}\fi

\bibitem[{Aditya et~al.(2015)Aditya, Yang, Baral, Fermuller, and
  Aloimonos}]{aditya2015images}
Somak Aditya, Yezhou Yang, Chitta Baral, Cornelia Fermuller, and Yiannis
  Aloimonos. 2015.
\newblock From images to sentences through scene description graphs using
  commonsense reasoning and knowledge.
\newblock \emph{arXiv preprint arXiv:1511.03292}.

\bibitem[{Agarwal et~al.(2021)Agarwal, Krueger, Clark, Radford, Kim, and
  Brundage}]{agarwal2021evaluating}
Sandhini Agarwal, Gretchen Krueger, Jack Clark, Alec Radford, Jong~Wook Kim,
  and Miles Brundage. 2021.
\newblock Evaluating {CLIP}: Towards characterization of broader capabilities
  and downstream implications.
\newblock \emph{arXiv preprint arXiv:2108.02818}.

\bibitem[{Anderson et~al.(2016)Anderson, Fernando, Johnson, and
  Gould}]{anderson2016spice}
Peter Anderson, Basura Fernando, Mark Johnson, and Stephen Gould. 2016.
\newblock Spice: Semantic propositional image caption evaluation.
\newblock In \emph{ECCV}. Springer.

\bibitem[{Artetxe and Schwenk(2019)}]{artetxe2019massively}
Mikel Artetxe and Holger Schwenk. 2019.
\newblock Massively multilingual sentence embeddings for zero-shot
  cross-lingual transfer and beyond.
\newblock \emph{TACL}, 7:597--610.

\bibitem[{Banerjee and Lavie(2005)}]{banerjee2005meteor}
Satanjeev Banerjee and Alon Lavie. 2005.
\newblock {METEOR:} an automatic metric for mt evaluation with improved
  correlation with human judgments.
\newblock In \emph{ACL workshop on Evaluation Measures for MT and
  Summarization}.

\bibitem[{Berg et~al.(2012)Berg, Berg, III, Dodge, Goyal, Han, Mensch,
  Mitchell, Sood, Stratos, and Yamaguchi}]{berg2012understanding}
Alexander~C. Berg, Tamara~L. Berg, Hal~Daumé III, Jesse Dodge, Amit Goyal,
  Xufeng Han, Alyssa Mensch, Margaret Mitchell, Aneesh Sood, Karl Stratos, and
  Kota Yamaguchi. 2012.
\newblock Understanding and predicting importance in images.
\newblock In \emph{CVPR}.

\bibitem[{Biten et~al.(2019)Biten, Gomez, Rusinol, and
  Karatzas}]{biten2019good}
Ali~Furkan Biten, Lluis Gomez, Mar{\c{c}}al Rusinol, and Dimosthenis Karatzas.
  2019.
\newblock Good news, everyone! context driven entity-aware captioning for news
  images.
\newblock In \emph{CVPR}.

\bibitem[{Blatz et~al.(2004)Blatz, Fitzgerald, Foster, Gandrabur, Goutte,
  Kulesza, Sanchis, and Ueffing}]{blatz2004confidence}
John Blatz, Erin Fitzgerald, George Foster, Simona Gandrabur, Cyril Goutte,
  Alex Kulesza, Alberto Sanchis, and Nicola Ueffing. 2004.
\newblock Confidence estimation for machine translation.
\newblock In \emph{COLING}.

\bibitem[{Chen et~al.(2020)Chen, Li, Yu, Kholy, Ahmed, Gan, Cheng, and
  Liu}]{chen2020uniter}
Yen-Chun Chen, Linjie Li, Licheng Yu, Ahmed~El Kholy, Faisal Ahmed, Zhe Gan,
  Yu~Cheng, and Jingjing Liu. 2020.
\newblock Uniter: Universal image-text representation learning.
\newblock In \emph{ECCV}.

\bibitem[{Cui et~al.(2018)Cui, Yang, Veit, Huang, and
  Belongie}]{cui2018learning}
Yin Cui, Guandao Yang, Andreas Veit, Xun Huang, and Serge Belongie. 2018.
\newblock Learning to evaluate image captioning.
\newblock In \emph{CVPR}.

\bibitem[{Dai and Lin(2017)}]{dai2017contrastive}
Bo~Dai and Dahua Lin. 2017.
\newblock Contrastive learning for image captioning.
\newblock In \emph{NeurIPS}.

\bibitem[{Devlin et~al.(2019)Devlin, Chang, Lee, and
  Toutanova}]{devlin2018bert}
Jacob Devlin, Ming-Wei Chang, Kenton Lee, and Kristina Toutanova. 2019.
\newblock {BERT:} pre-training of deep bidirectional transformers for language
  understanding.
\newblock In \emph{NAACL}.

\bibitem[{Dosovitskiy et~al.(2021)Dosovitskiy, Beyer, Kolesnikov, Weissenborn,
  Zhai, Unterthiner, Dehghani, Minderer, Heigold, Gelly, Uszkoreit, and
  Houlsby}]{dosovitskiy2020image}
Alexey Dosovitskiy, Lucas Beyer, Alexander Kolesnikov, Dirk Weissenborn,
  Xiaohua Zhai, Thomas Unterthiner, Mostafa Dehghani, Matthias Minderer, Georg
  Heigold, Sylvain Gelly, Jakob Uszkoreit, and Neil Houlsby. 2021.
\newblock An image is worth 16x16 words: Transformers for image recognition at
  scale.
\newblock In \emph{ICLR}.

\bibitem[{Elliott and Keller(2014)}]{elliott2014comparing}
Desmond Elliott and Frank Keller. 2014.
\newblock Comparing automatic evaluation measures for image description.
\newblock In \emph{ACL}.

\bibitem[{Gleason et~al.(2019)Gleason, Carrington, Cassidy, Morris, Kitani, and
  Bigham}]{gleason2019s}
Cole Gleason, Patrick Carrington, Cameron Cassidy, Meredith~Ringel Morris,
  Kris~M Kitani, and Jeffrey~P Bigham. 2019.
\newblock ``it's almost like they're trying to hide it": How user-provided
  image descriptions have failed to make twitter accessible.
\newblock In \emph{WWW}.

\bibitem[{Gleason et~al.(2020)Gleason, Pavel, McCamey, Low, Carrington, Kitani,
  and Bigham}]{gleason2020twitter}
Cole Gleason, Amy Pavel, Emma McCamey, Christina Low, Patrick Carrington,
  Kris~M Kitani, and Jeffrey~P Bigham. 2020.
\newblock Twitter a11y: A browser extension to make twitter images accessible.
\newblock In \emph{CHI}.

\bibitem[{Hendricks et~al.(2018)Hendricks, Burns, Saenko, Darrell, and
  Rohrbach}]{hendricks2018women}
Lisa~Anne Hendricks, Kaylee Burns, Kate Saenko, Trevor Darrell, and Anna
  Rohrbach. 2018.
\newblock Women also snowboard: Overcoming bias in captioning models.
\newblock In \emph{Proceedings of the European Conference on Computer Vision
  (ECCV)}, pages 771--787.

\bibitem[{Hodosh et~al.(2013)Hodosh, Young, and
  Hockenmaier}]{hodosh2013framing}
Micah Hodosh, Peter Young, and Julia Hockenmaier. 2013.
\newblock Framing image description as a ranking task: Data, models and
  evaluation metrics.
\newblock \emph{JAIR}, 47:853--899.

\bibitem[{Jia et~al.(2021)Jia, Yang, Xia, Chen, Parekh, Pham, Le, Sung, Li, and
  Duerig}]{jia2021scaling}
Chao Jia, Yinfei Yang, Ye~Xia, Yi-Ting Chen, Zarana Parekh, Hieu Pham, Quoc~V
  Le, Yunhsuan Sung, Zhen Li, and Tom Duerig. 2021.
\newblock Scaling up visual and vision-language representation learning with
  noisy text supervision.
\newblock In \emph{ICML}.

\bibitem[{Jiang et~al.(2019)Jiang, Huang, Zhang, Wang, Zhang, Gan, Diesner, and
  Gao}]{jiang2019tiger}
Ming Jiang, Qiuyuan Huang, Lei Zhang, Xin Wang, Pengchuan Zhang, Zhe Gan, Jana
  Diesner, and Jianfeng Gao. 2019.
\newblock {TIGEr:} text-to-image grounding for image caption evaluation.
\newblock In \emph{EMNLP}.

\bibitem[{K et~al.(2020)K, Wang, Mayhew, and Roth}]{wang2019cross}
Karthikeyan K, Zihan Wang, Stephen Mayhew, and Dan Roth. 2020.
\newblock Cross-lingual ability of multilingual {BERT}: An empirical study.
\newblock In \emph{ICLR}.

\bibitem[{Kane et~al.(2020)Kane, Kocyigit, Abdalla, Ajanoh, and
  Coulibali}]{kane2020nubia}
Hassan Kane, Muhammed~Yusuf Kocyigit, Ali Abdalla, Pelkins Ajanoh, and Mohamed
  Coulibali. 2020.
\newblock {NUBIA}: {N}e{U}ral based interchangeability assessor for text
  generation.
\newblock In \emph{1st Workshop on Evaluating NLG Evaluation}.

\bibitem[{Kilickaya et~al.(2017)Kilickaya, Erdem, Ikizler-Cinbis, and
  Erdem}]{kilickaya2016re}
Mert Kilickaya, Aykut Erdem, Nazli Ikizler-Cinbis, and Erkut Erdem. 2017.
\newblock Re-evaluating automatic metrics for image captioning.
\newblock In \emph{EACL}.

\bibitem[{Lee et~al.(2021)Lee, Yoon, Dernoncourt, Bui, and Jung}]{lee2021umic}
Hwanhee Lee, Seunghyun Yoon, Franck Dernoncourt, Trung Bui, and Kyomin Jung.
  2021.
\newblock {UMIC:} an unreferenced metric for image captioning via contrastive
  learning.
\newblock In \emph{ACL}.

\bibitem[{Lee et~al.(2020)Lee, Yoon, Dernoncourt, Kim, Bui, and
  Jung}]{lee2020vilbertscore}
Hwanhee Lee, Seunghyun Yoon, Franck Dernoncourt, Doo~Soon Kim, Trung Bui, and
  Kyomin Jung. 2020.
\newblock Vilbertscore: Evaluating image caption using vision-and-language
  bert.
\newblock In \emph{First Workshop on Evaluation and Comparison of NLP Systems}.

\bibitem[{Lee et~al.(2018)Lee, Chen, Hua, Hu, and He}]{lee2018stacked}
Kuang-Huei Lee, Xi~Chen, Gang Hua, Houdong Hu, and Xiaodong He. 2018.
\newblock Stacked cross attention for image-text matching.
\newblock In \emph{ECCV}.

\bibitem[{Lin(2004)}]{lin2004rouge}
Chin-Yew Lin. 2004.
\newblock Rouge: A package for automatic evaluation of summaries.
\newblock \emph{Text Summarization Branches Out}.

\bibitem[{Lin et~al.(2014)Lin, Maire, Belongie, Hays, Perona, Ramanan,
  Doll{\'a}r, and Zitnick}]{lin2014microsoft}
Tsung-Yi Lin, Michael Maire, Serge Belongie, James Hays, Pietro Perona, Deva
  Ramanan, Piotr Doll{\'a}r, and C~Lawrence Zitnick. 2014.
\newblock Microsoft {COCO}: Common objects in context.
\newblock In \emph{ECCV}. Springer.

\bibitem[{Liu et~al.(2018)Liu, Li, Shao, Chen, and Wang}]{liu2018show}
Xihui Liu, Hongsheng Li, Jing Shao, Dapeng Chen, and Xiaogang Wang. 2018.
\newblock Show, tell and discriminate: Image captioning by self-retrieval with
  partially labeled data.
\newblock In \emph{ECCV}.

\bibitem[{Lo(2019)}]{lo2019yisi}
Chi-kiu Lo. 2019.
\newblock Yisi-a unified semantic mt quality evaluation and estimation metric
  for languages with different levels of available resources.
\newblock In \emph{Fourth Conference on Machine Translation}.

\bibitem[{Louis and Nenkova(2013)}]{louis2013automatically}
Annie Louis and Ani Nenkova. 2013.
\newblock Automatically assessing machine summary content without a gold
  standard.
\newblock \emph{Computational Linguistics}, 39(2):267--300.

\bibitem[{Lu et~al.(2019)Lu, Batra, Parikh, and Lee}]{lu2019vilbert}
Jiasen Lu, Dhruv Batra, Devi Parikh, and Stefan Lee. 2019.
\newblock {ViLBERT}: Pretraining task-agnostic visiolinguistic representations
  for vision-and-language tasks.
\newblock In \emph{NeurIPS}.

\bibitem[{Lu et~al.(2020)Lu, Goswami, Rohrbach, Parikh, and Lee}]{lu202012}
Jiasen Lu, Vedanuj Goswami, Marcus Rohrbach, Devi Parikh, and Stefan Lee. 2020.
\newblock 12-in-1: Multi-task vision and language representation learning.
\newblock In \emph{CVPR}.

\bibitem[{Luo et~al.(2021)Luo, Darrell, and Rohrbach}]{luo2021newsclippings}
Grace Luo, Trevor Darrell, and Anna Rohrbach. 2021.
\newblock {NewsCLIPpings:} automatic generation of out-of-context multimodal
  media.
\newblock \emph{arXiv preprint arXiv:2104.05893}.

\bibitem[{Luo et~al.(2018)Luo, Price, Cohen, and
  Shakhnarovich}]{luo2018discriminability}
Ruotian Luo, Brian Price, Scott Cohen, and Gregory Shakhnarovich. 2018.
\newblock Discriminability objective for training descriptive captions.
\newblock In \emph{CVPR}.

\bibitem[{MacLeod et~al.(2017)MacLeod, Bennett, Morris, and
  Cutrell}]{macleod2017understanding}
Haley MacLeod, Cynthia~L Bennett, Meredith~Ringel Morris, and Edward Cutrell.
  2017.
\newblock Understanding blind people's experiences with computer-generated
  captions of social media images.
\newblock In \emph{CHI}.

\bibitem[{Madhyastha et~al.(2019)Madhyastha, Wang, and
  Specia}]{madhyastha-etal-2019-vifidel}
Pranava Madhyastha, Josiah Wang, and Lucia Specia. 2019.
\newblock {VIFIDEL}: Evaluating the visual fidelity of image descriptions.
\newblock In \emph{ACL}.

\bibitem[{Mehdad et~al.(2012)Mehdad, Negri, and Federico}]{mehdad2012match}
Yashar Mehdad, Matteo Negri, and Marcello Federico. 2012.
\newblock Match without a referee: evaluating mt adequacy without reference
  translations.
\newblock In \emph{Seventh Workshop on Statistical Machine Translation}.

\bibitem[{Mehri and Eskenazi(2020)}]{mehri2020usr}
Shikib Mehri and Maxine Eskenazi. 2020.
\newblock {USR}: An unsupervised and reference free evaluation metric for
  dialog generation.
\newblock In \emph{ACL}.

\bibitem[{Mitchell et~al.(2019)Mitchell, Wu, Zaldivar, Barnes, Vasserman,
  Hutchinson, Spitzer, Raji, and Gebru}]{mitchell2019model}
Margaret Mitchell, Simone Wu, Andrew Zaldivar, Parker Barnes, Lucy Vasserman,
  Ben Hutchinson, Elena Spitzer, Inioluwa~Deborah Raji, and Timnit Gebru. 2019.
\newblock Model cards for model reporting.
\newblock In \emph{FAccT}.

\bibitem[{Oord et~al.(2018)Oord, Li, and Vinyals}]{oord2018representation}
Aaron van~den Oord, Yazhe Li, and Oriol Vinyals. 2018.
\newblock Representation learning with contrastive predictive coding.
\newblock \emph{arXiv preprint arXiv:1807.03748}.

\bibitem[{Papineni et~al.(2002)Papineni, Roukos, Ward, and
  Zhu}]{papineni2002bleu}
Kishore Papineni, Salim Roukos, Todd Ward, and Wei-Jing Zhu. 2002.
\newblock Bleu: a method for automatic evaluation of machine translation.
\newblock In \emph{ACL}.

\bibitem[{Pedregosa et~al.(2011)Pedregosa, Varoquaux, Gramfort, Michel,
  Thirion, Grisel, Blondel, Prettenhofer, Weiss, Dubourg, Vanderplas, Passos,
  Cournapeau, Brucher, Perrot, and Duchesnay}]{scikit-learn}
F.~Pedregosa, G.~Varoquaux, A.~Gramfort, V.~Michel, B.~Thirion, O.~Grisel,
  M.~Blondel, P.~Prettenhofer, R.~Weiss, V.~Dubourg, J.~Vanderplas, A.~Passos,
  D.~Cournapeau, M.~Brucher, M.~Perrot, and E.~Duchesnay. 2011.
\newblock Scikit-learn: Machine learning in {P}ython.
\newblock \emph{JMLR}, 12.

\bibitem[{Peyrard and Gurevych(2018)}]{peyrard2018objective}
Maxime Peyrard and Iryna Gurevych. 2018.
\newblock Objective function learning to match human judgements for
  optimization-based summarization.
\newblock In \emph{NAACL}.

\bibitem[{Pires et~al.(2019)Pires, Schlinger, and
  Garrette}]{pires2019multilingual}
Telmo Pires, Eva Schlinger, and Dan Garrette. 2019.
\newblock How multilingual is multilingual {BERT}?
\newblock In \emph{ACL}.

\bibitem[{Radford et~al.(2021)Radford, Kim, Hallacy, Ramesh, Goh, Agarwal,
  Sastry, Askell, Mishkin, Clark, Krueger, and Sutskever}]{radford2learning}
Alec Radford, Jong~Wook Kim, Chris Hallacy, Aditya Ramesh, Gabriel Goh,
  Sandhini Agarwal, Girish Sastry, Amanda Askell, Pamela Mishkin, Jack Clark,
  Gretchen Krueger, and Ilya Sutskever. 2021.
\newblock Learning transferable visual models from natural language
  supervision.

\bibitem[{Radford et~al.(2019)Radford, Wu, Child, Luan, Amodei, and
  Sutskever}]{radford2019language}
Alec Radford, Jeffrey Wu, Rewon Child, David Luan, Dario Amodei, and Ilya
  Sutskever. 2019.
\newblock Language models are unsupervised multitask learners.
\newblock \emph{OpenAI blog}, 1(8):9.

\bibitem[{Rohrbach et~al.(2018)Rohrbach, Hendricks, Burns, Darrell, and
  Saenko}]{rohrbach2018object}
Anna Rohrbach, Lisa~Anne Hendricks, Kaylee Burns, Trevor Darrell, and Kate
  Saenko. 2018.
\newblock Object hallucination in image captioning.
\newblock In \emph{EMNLP}.

\bibitem[{Rohrbach et~al.(2017)Rohrbach, Torabi, Rohrbach, Tandon, Pal,
  Larochelle, Courville, and Schiele}]{rohrbach2017movie}
Anna Rohrbach, Atousa Torabi, Marcus Rohrbach, Niket Tandon, Christopher Pal,
  Hugo Larochelle, Aaron Courville, and Bernt Schiele. 2017.
\newblock Movie description.
\newblock \emph{IJCV}.

\bibitem[{Sennrich et~al.(2016)Sennrich, Haddow, and
  Birch}]{sennrich2015neural}
Rico Sennrich, Barry Haddow, and Alexandra Birch. 2016.
\newblock Neural machine translation of rare words with subword units.
\newblock In \emph{ACL}.

\bibitem[{Shekhar et~al.(2017)Shekhar, Pezzelle, Klimovich, Herbelot, Nabi,
  Sangineto, and Bernardi}]{shekhar2017foil}
Ravi Shekhar, Sandro Pezzelle, Yauhen Klimovich, Aur{\'e}lie Herbelot, Moin
  Nabi, Enver Sangineto, and Raffaella Bernardi. 2017.
\newblock {FOIL} it! find one mismatch between image and language caption.
\newblock In \emph{ACL}.

\bibitem[{Shuster et~al.(2019)Shuster, Humeau, Hu, Bordes, and
  Weston}]{shuster2019engaging}
Kurt Shuster, Samuel Humeau, Hexiang Hu, Antoine Bordes, and Jason Weston.
  2019.
\newblock Engaging image captioning via personality.
\newblock In \emph{CVPR}.

\bibitem[{Sohn(2016)}]{sohn2016improved}
Kihyuk Sohn. 2016.
\newblock Improved deep metric learning with multi-class n-pair loss objective.
\newblock In \emph{NeurIPS}.

\bibitem[{Specia et~al.(2010)Specia, Raj, and Turchi}]{specia2010machine}
Lucia Specia, Dhwaj Raj, and Marco Turchi. 2010.
\newblock Machine translation evaluation versus quality estimation.
\newblock \emph{Machine translation}, 24(1):39--50.

\bibitem[{Specia and Shah(2018)}]{specia2018machine}
Lucia Specia and Kashif Shah. 2018.
\newblock Machine translation quality estimation: Applications and future
  perspectives.
\newblock In \emph{Translation Quality Assessment}, pages 201--235. Springer.

\bibitem[{Stangl et~al.(2020)Stangl, Morris, and Gurari}]{stangl2020person}
Abigale Stangl, Meredith~Ringel Morris, and Danna Gurari. 2020.
\newblock ``person, shoes, tree. is the person naked?" what people with vision
  impairments want in image descriptions.
\newblock In \emph{CHI}.

\bibitem[{Sun and Nenkova(2019)}]{sun2019feasibility}
Simeng Sun and Ani Nenkova. 2019.
\newblock The feasibility of embedding based automatic evaluation for single
  document summarization.
\newblock In \emph{EMNLP}.

\bibitem[{Tao et~al.(2018)Tao, Mou, Zhao, and Yan}]{tao2018ruber}
Chongyang Tao, Lili Mou, Dongyan Zhao, and Rui Yan. 2018.
\newblock Ruber: An unsupervised method for automatic evaluation of open-domain
  dialog systems.
\newblock In \emph{AAAI}.

\bibitem[{Vaswani et~al.(2017)Vaswani, Shazeer, Parmar, Uszkoreit, Jones,
  Gomez, Kaiser, and Polosukhin}]{vaswani2017attention}
Ashish Vaswani, Noam Shazeer, Niki Parmar, Jakob Uszkoreit, Llion Jones,
  Aidan~N Gomez, Lukasz Kaiser, and Illia Polosukhin. 2017.
\newblock Attention is all you need.
\newblock In \emph{NeurIPS}.

\bibitem[{Vedantam et~al.(2015)Vedantam, Lawrence~Zitnick, and
  Parikh}]{vedantam2015cider}
Ramakrishna Vedantam, C~Lawrence~Zitnick, and Devi Parikh. 2015.
\newblock Cider: Consensus-based image description evaluation.
\newblock In \emph{CVPR}.

\bibitem[{Vinyals et~al.(2016)Vinyals, Toshev, Bengio, and
  Erhan}]{vinyals2016show}
Oriol Vinyals, Alexander Toshev, Samy Bengio, and Dumitru Erhan. 2016.
\newblock Show and tell: Lessons learned from the 2015 mscoco image captioning
  challenge.
\newblock \emph{TPAMI}, 39(4):652--663.

\bibitem[{Wang et~al.(2021)Wang, Yao, Wang, Wu, and Chen}]{wang2021faier}
Sijin Wang, Ziwei Yao, Ruiping Wang, Zhongqin Wu, and Xilin Chen. 2021.
\newblock {FAIEr}: Fidelity and adequacy ensured image caption evaluation.
\newblock In \emph{CVPR}.

\bibitem[{Wu and Dredze(2019)}]{Wu:2019rw}
Shijie Wu and Mark Dredze. 2019.
\newblock Beto, bentz, becas: The surprising cross-lingual effectiveness of
  {BERT}.
\newblock In \emph{EMNLP}.

\bibitem[{Yankovskaya et~al.(2019)Yankovskaya, T{\"a}ttar, and
  Fishel}]{yankovskaya2019quality}
Elizaveta Yankovskaya, Andre T{\"a}ttar, and Mark Fishel. 2019.
\newblock Quality estimation and translation metrics via pre-trained word and
  sentence embeddings.
\newblock In \emph{Fourth Conference on Machine Translation}.

\bibitem[{Yi et~al.(2020)Yi, Deng, and Hu}]{yi2020improving}
Yanzhi Yi, Hangyu Deng, and Jinglu Hu. 2020.
\newblock Improving image captioning evaluation by considering inter references
  variance.
\newblock In \emph{ACL}.

\bibitem[{Young et~al.(2014)Young, Lai, Hodosh, and
  Hockenmaier}]{young2014image}
Peter Young, Alice Lai, Micah Hodosh, and Julia Hockenmaier. 2014.
\newblock From image descriptions to visual denotations: New similarity metrics
  for semantic inference over event descriptions.
\newblock \emph{TACL}, 2:67--78.

\bibitem[{Zhang et~al.(2020)Zhang, Kishore, Wu, Weinberger, and
  Artzi}]{zhang2019bertscore}
Tianyi Zhang, Varsha Kishore, Felix Wu, Kilian~Q Weinberger, and Yoav Artzi.
  2020.
\newblock {BERTScore}: Evaluating text generation with {BERT}.
\newblock In \emph{ICLR}.

\bibitem[{Zhao et~al.(2020)Zhao, Glava{\v{s}}, Peyrard, Gao, West, and
  Eger}]{zhao-etal-2020-limitations}
Wei Zhao, Goran Glava{\v{s}}, Maxime Peyrard, Yang Gao, Robert West, and
  Steffen Eger. 2020.
\newblock On the limitations of cross-lingual encoders as exposed by
  reference-free machine translation evaluation.
\newblock In \emph{ACL}.

\bibitem[{Zitnick and Parikh(2013)}]{zitnick2013bringing}
C~Lawrence Zitnick and Devi Parikh. 2013.
\newblock Bringing semantics into focus using visual abstraction.
\newblock In \emph{CVPR}.

\end{thebibliography}


\begin{thebibliography}{5}
\expandafter\ifx\csname natexlab\endcsname\relax\def\natexlab#1{#1}\fi

\bibitem[{Anderson et~al.(2016)Anderson, Fernando, Johnson, and
  Gould}]{anderson2016spice}
Peter Anderson, Basura Fernando, Mark Johnson, and Stephen Gould. 2016.
\newblock Spice: Semantic propositional image caption evaluation.
\newblock In \emph{ECCV}. Springer.

\bibitem[{Card et~al.(2020)Card, Henderson, Khandelwal, Jia, Mahowald, and
  Jurafsky}]{card2020little}
Dallas Card, Peter Henderson, Urvashi Khandelwal, Robin Jia, Kyle Mahowald, and
  Dan Jurafsky. 2020.
\newblock With little power comes great responsibility.
\newblock In \emph{EMNLP}.

\bibitem[{Cui et~al.(2018)Cui, Yang, Veit, Huang, and
  Belongie}]{cui2018learning}
Yin Cui, Guandao Yang, Andreas Veit, Xun Huang, and Serge Belongie. 2018.
\newblock Learning to evaluate image captioning.
\newblock In \emph{CVPR}.

\bibitem[{Kane et~al.(2020)Kane, Kocyigit, Abdalla, Ajanoh, and
  Coulibali}]{kane2020nubia}
Hassan Kane, Muhammed~Yusuf Kocyigit, Ali Abdalla, Pelkins Ajanoh, and Mohamed
  Coulibali. 2020.
\newblock {NUBIA}: {N}e{U}ral based interchangeability assessor for text
  generation.
\newblock In \emph{1st Workshop on Evaluating NLG Evaluation}.

\bibitem[{Zhang et~al.(2020)Zhang, Kishore, Wu, Weinberger, and
  Artzi}]{zhang2019bertscore}
Tianyi Zhang, Varsha Kishore, Felix Wu, Kilian~Q Weinberger, and Yoav Artzi.
  2020.
\newblock {BERTScore}: Evaluating text generation with {BERT}.
\newblock In \emph{ICLR}.

\end{thebibliography}
\end{document}